\pdfoutput=1

\documentclass[11pt]{article}

\usepackage{acl}

\usepackage{times}
\usepackage{latexsym}

\usepackage[T1]{fontenc}

\usepackage[utf8]{inputenc}

\usepackage{microtype}

\usepackage{inconsolata}

\usepackage{graphicx}
\usepackage{enumerate}
\usepackage{enumitem}
\usepackage{array}
\usepackage{booktabs}
\usepackage{multirow}
\usepackage{fdsymbol}

\usepackage{amsmath,amsfonts,bm}

\def\eqref#1{equation~\ref{#1}}

\def\1{\bm{1}}

\newcommand{\ie}{{\em i.e.,}}
\newcommand{\eg}{{\em e.g.,}}
\newcommand{\Ni}{({\em i})~}
\newcommand{\Nii}{({\em ii})~}
\newcommand{\Niii}{({\em iii})~}

\newcommand{\Na}{({\em a})~}
\newcommand{\Nb}{({\em b})~}

\DeclareMathAlphabet{\mathsfit}{\encodingdefault}{\sfdefault}{m}{sl}
\SetMathAlphabet{\mathsfit}{bold}{\encodingdefault}{\sfdefault}{bx}{n}

\usepackage[nameinlink]{cleveref}
\crefformat{section}{\S#2#1#3} 
\crefname{algorithm}{Alg.}{Algs.}
\crefformat{subsection}{\S#2#1#3}
\Crefname{equation}{Eq.}{Eqs.}
\Crefname{figure}{Fig.}{Figs.}
\newcommand{\rateft}{RATE-FT}
\title{Learning Auxiliary Tasks Improves Reference-Free Hallucination Detection \\in Open-Domain Long-Form Generation}

\author{Chengwei Qin$^\dagger$$^\vardiamondsuit$\thanks{Correspondence to Chengwei Qin <qcwthu@gmail.com> and Karthik Abinav Sankararaman <karthikabinavs@gmail.com>}, Wenxuan Zhou$^\clubsuit$, Karthik Abinav Sankararaman$^\clubsuit$, \\ \textbf{Nanshu Wang$^\clubsuit$, Tengyu Xu$^\clubsuit$, Alexander Radovic$^\clubsuit$, Eryk Helenowski$^\clubsuit$}, \\ \textbf{Arya Talebzadeh$^\clubsuit$, Aditya Tayade$^\clubsuit$, Sinong Wang$^\clubsuit$, Shafiq Joty$^\vardiamondsuit$, Han Fang$^\clubsuit$, Hao Ma$^\clubsuit$}
\\
$^\dagger$HKUST(GZ), $^\vardiamondsuit$NTU, Singapore, $^\clubsuit$GenAI, Meta}

\begin{document}
\maketitle
\begin{abstract}
Hallucination, the generation of factually incorrect information, remains a significant challenge for large language models (LLMs), especially in open-domain long-form generation. Existing approaches for detecting hallucination in long-form tasks either focus on limited domains or rely heavily on external fact-checking tools, which may not always be available. 

In this work, we systematically investigate reference-free hallucination detection in open-domain long-form responses. Our findings reveal that internal states (\eg\ model's output probability and entropy) alone are insufficient for reliably (\ie\ better than random guessing) distinguishing between factual and hallucinated content. To enhance detection, we explore various existing approaches, including prompting-based methods, probing, and fine-tuning, with fine-tuning proving the most effective. To further improve the accuracy, we introduce a new paradigm, named RATE-FT, that augments fine-tuning with an auxiliary question answering task for the model to jointly learn with the main task of hallucination detection. With extensive experiments and analysis using a variety of model families and datasets, we demonstrate the effectiveness and generalizability of our method, \eg\ +3\% over general fine-tuning methods on LongFact.

\end{abstract}

\section{Introduction}

\begin{figure}[t]
  \centering
    \includegraphics[width=0.44\textwidth]{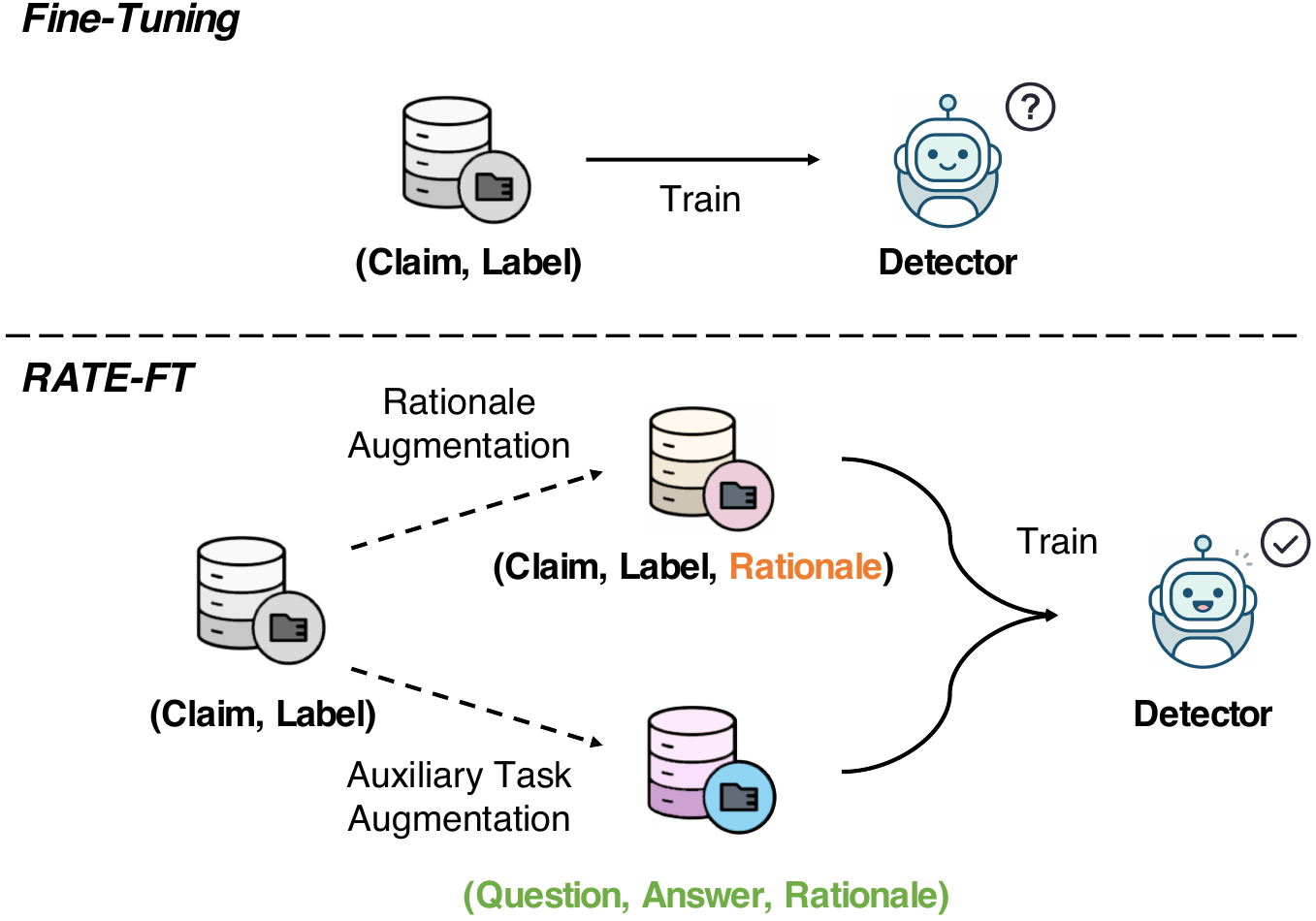}
 \captionof{figure}{Comparison between Fine-Tuning and \rateft\ for hallucination detection. \rateft\ improves Fine-Tuning by incorporating rationales and an auxiliary task (question answering) into the training process.}
  \label{fig:method_rateft}
  \vspace{-1.0em}
\end{figure}

With the recent advancements in model scale and pretraining data, large language models (LLMs) have demonstrated remarkable capabilities in various natural language processing (NLP) tasks \citep{brown2020language}. Despite these successes, hallucination, where models tend to produce content that conflicts with real-world facts, remains a significant challenge \citep{zhang2023siren}. Most existing research on hallucination detection has focused on short-form tasks, where the output consists of one or a few tokens. While these methods are effective for short-form content \citep{manakul-etal-2023-selfcheckgpt,mahaut-etal-2024-factual,yehuda-etal-2024-interrogatellm,zhang-etal-2024-transferable}, extending them to open-domain long-form generation presents additional complexities and new challenges. Unlike short-form tasks, long-form responses can span hundreds or even thousands of tokens, requiring models to generate detailed and nuanced answers to broad fact-seeking prompts \citep{wei2024long}. This necessitates that LLMs synthesize information across multiple knowledge domains, increasing the risk of generating content that sounds plausible yet is factually incorrect. For example, when answering `What is the significance of Amber Room?', LLMs may generate responses that mix accurate historical information with fabricated details, complicating the task of distinguishing fact from hallucination.

Recent efforts have sought to address hallucination detection in long-form tasks. However, they either focus on limited domains, \eg\ biography generation \citep{min-etal-2023-factscore,fadeeva-etal-2024-fact} or rely heavily on external fact-checking tools or knowledge bases, \eg\ Google Search 
\citep{wei2024long}. While these tools offer valuable support, they are not always available or scalable. This raises an important question: \emph{can we develop hallucination detectors that rely solely on the model itself, without the need for external fact-checking resources?} So far, little attention has been given to systematically exploring how the model's own mechanisms can be used for detecting hallucinations in open-domain long-form generation.

To address this gap, we start by investigating hallucination detection in open-domain long-form responses using the model's internal states, \eg\ output probability and entropy. Specifically, we decompose long-form responses into atomized claims using the model and verify each claim's correctness using Google Search to construct benchmark data following \citet{wei2024long}. Our analysis reveals that these internal states alone are insufficient for reliably (\ie\ better than random guessing) distinguishing between correct and incorrect claims, indicating that the mechanisms for detecting hallucinations in long-form outputs differ significantly from those in short-form tasks. To enhance detection, we explore several existing methods, including prompting, probing, and fine-tuning LLMs. Our experimental results show that fine-tuning LLMs is the most effective method to detect hallucinations. 

Building on this, we introduce a novel method Rationale and Auxiliary Task Enhanced Fine-Tuning (\rateft) (see Figure \ref{fig:method_rateft} for the comparison between Fine-Tuning and \rateft). Specifically, we convert the original claims into auxiliary question answering (QA) examples for augmentation, providing a complementary learning perspective for the model, which enables better generalization. Additionally, we incorporate collected rationales into the training process for better reasoning. Extensive experiments and analysis using different models demonstrate the effectiveness and generalizability of our approach. Furthermore, we investigate the integration of model uncertainty into hallucination detection. By allowing the model to classify claims as ``unknown'' when uncertain, we open the door for incorporating external tools when needed, enabling a more robust, hybrid detection pipeline (detailed in Section \ref{sec:uncertainty_for_detection}). In summary, our main contributions are:

\begin{itemize}[leftmargin=*,topsep=5pt,itemsep=5pt,parsep=2pt]
    \item To the best of our knowledge, we are the first to systematically investigate reference-free hallucination detection in open-domain long-form generation by analyzing a representative set of existing methods.
    \item We introduce a novel approach (\rateft) that incorporates rationales and an auxiliary question answering task into fine-tuning. Through extensive experiments and analysis, we demonstrate the effectiveness of \rateft\ across two benchmark datasets.
\end{itemize}

\begin{figure*}[t]
  \centering
  \begin{minipage}[ht]{0.48\textwidth}
  \centering
    \includegraphics[width=1.0\textwidth]{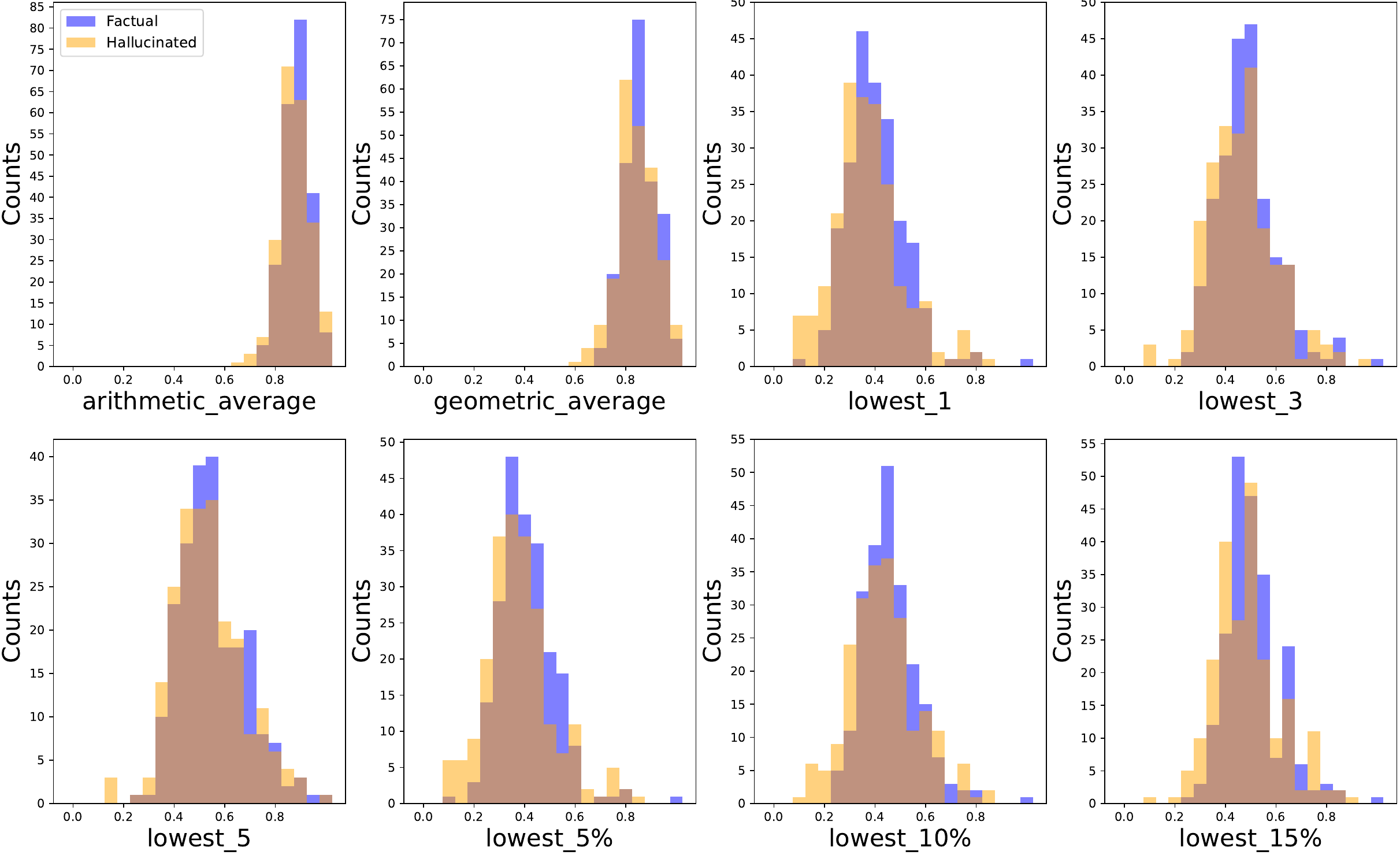}
 \captionof{figure}{Hallucination detection results based on token probability.}
  \label{fig:token_probability}
  \end{minipage}
  \hfill
  \begin{minipage}[ht]{0.475\textwidth}
  \centering
  \includegraphics[width=1.0\textwidth]{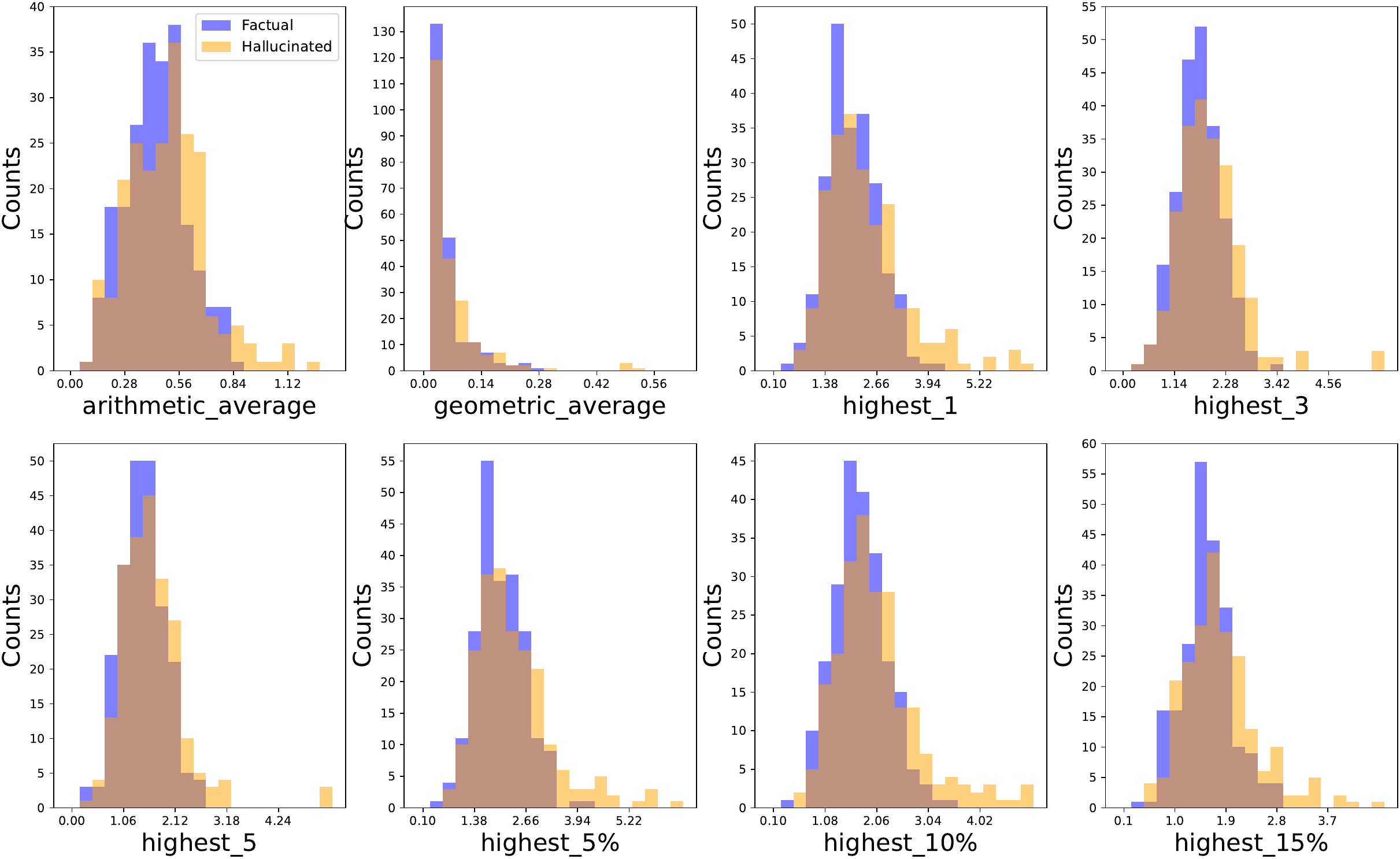}
    \captionof{figure}{Hallucination detection results based on token entropy (uncertainty).}
    \label{fig:token_entropy}
\end{minipage}
\vspace{-0.1em}
\end{figure*}

\section{Related Work} \label{sec:related_work}

Large Language Models (LLMs) often generate content that appears plausible but is factually unsupported, a phenomenon commonly referred to as hallucination \citep{zhang2023siren}. Hallucinations are typically categorized based on what they contradict: factuality hallucinations (the main focus of this work), which conflict with real-world knowledge, and faithfulness hallucinations, which contradict the input context, instruction, or internal logic \citep{huang2024survey}. These two types form the foundation for much of the recent work on LLM hallucination. 

An increasing number of studies have investigated the underlying causes of hallucination in LLMs, which can arise across the data, training, and inference aspects. At the data level, hallucinations often stem from flawed sources, such as misinformation, social biases and limited knowledge coverage \citep{hernandez2022scaling,ladhak-etal-2023-pre,li-etal-2023-large}, as well as from inferior data utilization, including the model's tendency to rely on knowledge shortcuts \citep{pmlr-v202-kandpal23a,kang-choi-2023-impact,liu-etal-2024-lost}. During training, factors like exposure bias and side effects of reinforcement learning from human feedback (RLHF), \eg\ sycophantic responses, can further amplify hallucination tendencies \citep{wei2023simple,zhang2024how}. At inference time, hallucinations may result from insufficient context attention and the inherent randomness of decoding strategies \citep{chen2023improving,chuang2024dola,shi-etal-2024-trusting}. These diverse and interdependent causes underscore the importance of addressing hallucination through a holistic perspective.

In response, various hallucination detection methods has been proposed, typically categorized by the type of hallucination: factuality and faithfulness. Factuality detection assesses whether generated content aligns with real-world facts, leveraging techniques such as external retrieval-based fact-checking and internal uncertainty estimation \citep{min-etal-2023-factscore,zhao2023felm,manakul-etal-2023-selfcheckgpt,chen-etal-2024-complex,fadeeva-etal-2024-fact,wei2024long}. Faithfulness detection, on the other hand, focuses on whether outputs remain consistent with the input context or instructions. Common approaches include fact overlap metrics, natural language inference (NLI) classifiers, and LLM-based evaluation via prompting \citep{nan-etal-2021-entity,laban-etal-2022-summac,laban2023llms,adlakha-etal-2024-evaluating}. To mitigate hallucinations, researchers have developed different strategies for each stage. Data-level hallucinations are addressed through data curation, model editing for knowledge updates, and retrieval-augmented generation (RAG) \citep{shao-etal-2023-enhancing,meng2023massediting,gunasekar2024textbooks}. Training-related hallucinations are mitigated by refining pre-training objectives, reducing exposure bias, and minimizing misalignment introduced by supervised fine-tuning (SFT) and RLHF \citep{bertsch-etal-2023-mbr,sharma2024towards,shi2024incontext}. For inference-time hallucinations, improved decoding strategies have been explored to enhance both factuality and faithfulness \citep{li2023inferencetime,wan-etal-2023-faithfulness,shi-etal-2024-trusting}.

However, most existing hallucination detection methods have primarily focused on short-form tasks, where the output consists of one or a few tokens. In this work, we shift the focus to the more challenging problem of reference-free hallucination detection in open-domain long-form generation, where outputs are substantially longer and require a more nuanced evaluation of factuality.

\section{Are LLMs' Internal States Sufficient for Open-Domain Long-Form Generation?} \label{sec:internal}

The internal states of LLMs, such as output probability and entropy, have been shown to be effective in detecting hallucinations in short-form tasks, where outputs are typically limited to only a few tokens. By analyzing these signals, models can often differentiate between factual and hallucinated information. However, their applicability in open-domain long-form generation remains underexplored. A key question is whether LLMs can depend solely on their internal states to identify hallucinations in long-form generation, without using external fact-checking tools. 

To answer it, we conduct some pilot experiments on LongFact \citep{wei2024long}, a long-form generation dataset spanning 38 different domains. Specifically, for each prompt in the sampled subset (200 prompts), we obtain a long-form response from Llama-3-8B-Instruct with greedy decoding. Following \citet{wei2024long}, we employ the model to decompose long-form responses into atomized claims and assess whether each claim is relevant to answering the corresponding prompt. For each relevant claim, we use the model to generate multi-step Google Search queries and reason about whether the search results support the claim. Claims supported by the search results are labeled as ``factual'', while those contradicted by the results are categorized as ``hallucinated''. After construction, we obtain 2394 factual claims and 223 hallucinated claims, respectively. We then randomly selected an equal number (223) of factual and hallucinated claims for experiments.

\begin{figure*}[t]
  \centering
  \begin{minipage}[ht]{0.48\textwidth}
  \centering
    \includegraphics[width=1.0\textwidth]{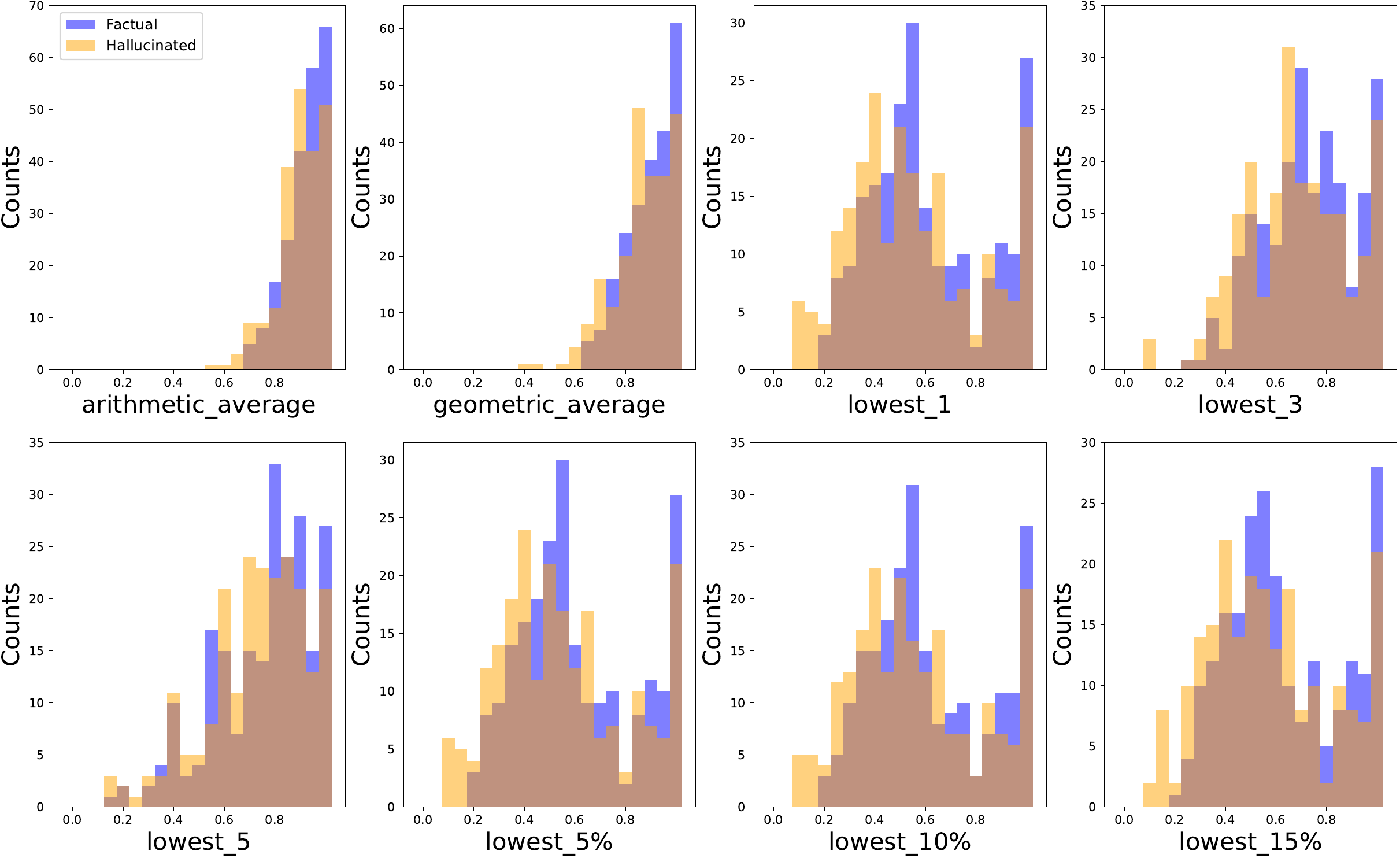}
 \captionof{figure}{Hallucination detection results based on the probability of entity-related tokens.}
  \label{fig:token_probability_entity}
  \end{minipage}
  \hfill
  \begin{minipage}[ht]{0.48\textwidth}
  \centering
  \includegraphics[width=1.0\textwidth]{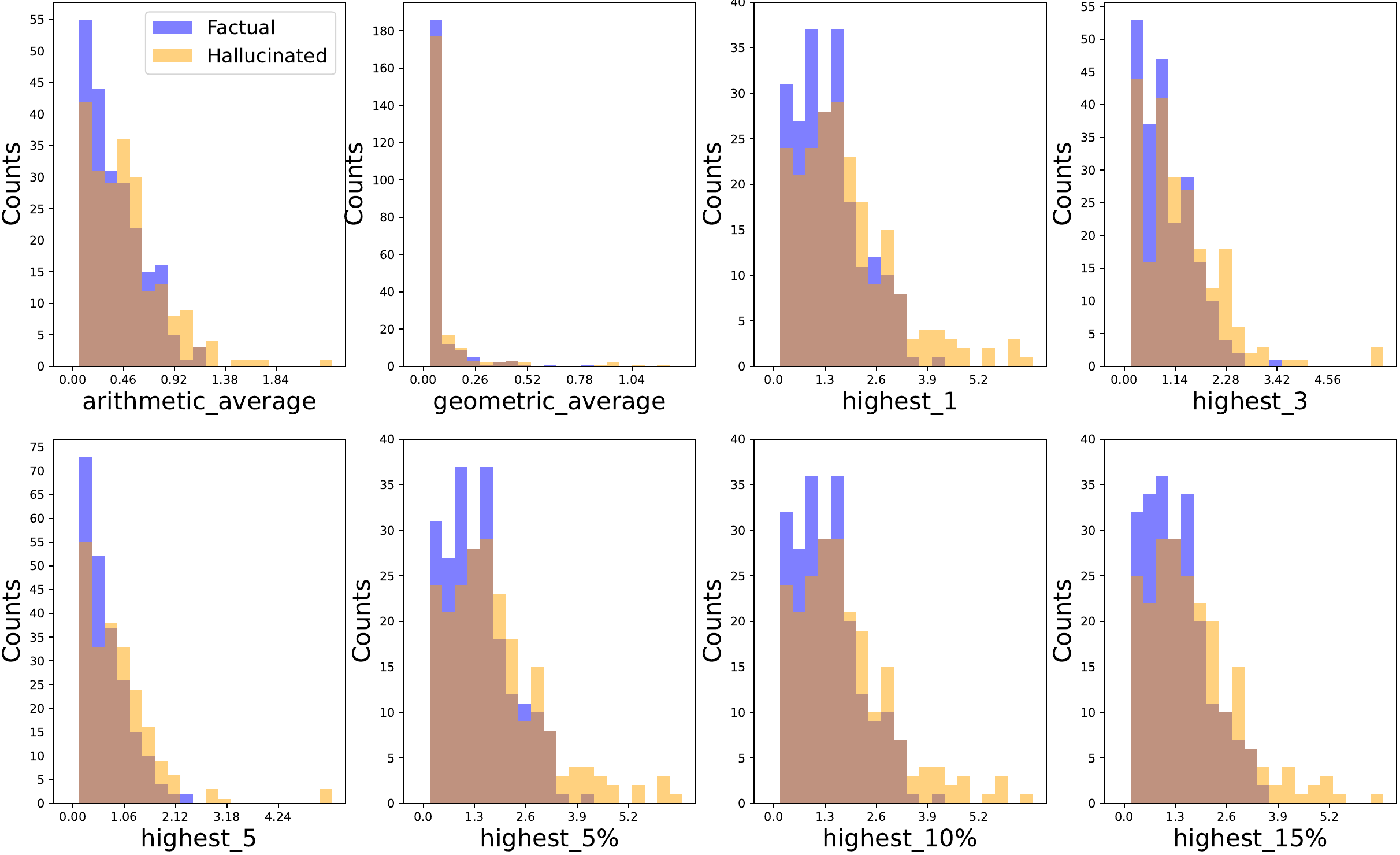}
    \captionof{figure}{Hallucination detection results based on the entropy of entity-related tokens.}
    \label{fig:token_entropy_entity}
  \end{minipage}
\end{figure*}

For each claim, we mainly focus on two types of internal states to estimate factual confidence following SelfCheckGPT \citep{manakul-etal-2023-selfcheckgpt}: the probability or the entropy (uncertainty) of output tokens. Specifically, we examine the arithmetic and geometric \footnote{The geometric average of all tokens is commonly known as perplexity.} averages of all tokens, the average of tokens with the top-$K$ lowest probability or highest entropy ($K = 1, 3, 5$), and the average of tokens with the top-$P\%$ lowest probability or highest entropy ($P = 5, 10, 15$). The results in Figure \ref{fig:token_probability} and \ref{fig:token_entropy} suggest that neither internal state reliably, \ie\ better than random guessing, predicts the correctness of a given claim, which may be due to the presence of numerous insignificant tokens within the claim, such as stop words. To address this, we consider variants that focus only on output tokens related to entities. The results, shown in Figure \ref{fig:token_probability_entity} and \ref{fig:token_entropy_entity}, reveal similar patterns. We analyze the underlying reasons as follows. In open-domain long-form generation, claims are not limited to a few tokens, which introduces multiple sources of uncertainty. Specifically, \emph{the probability or entropy reflects the model's confidence in how a claim is expressed,} \ie\ \emph{its confidence in the claim as a sequence of output tokens, rather than in the correctness of the claim}. Different surface forms of the claim yield different confidence levels, leading to unreliable estimates. Note that our analysis differs from that of SelfCheckGPT \citep{manakul-etal-2023-selfcheckgpt} in the following key aspects.

\begin{itemize}[leftmargin=*,topsep=5pt,itemsep=5pt,parsep=2pt]
    \item SelfCheckGPT only examines the arithmetic average of all tokens and the average of tokens with the top-1 lowest probability or highest entropy. In contrast, our work covers a broader range of variants. 
    \item Our findings differ significantly from those reported in SelfCheckGPT. While SelfCheckGPT suggests that LLM probabilities correlate well with factuality, our experiments demonstrate that neither internal state reliably, \ie\ better than random guessing, predicts the correctness of a given claim. Importantly, our findings are consistent with those in \citet{kapoor2024large}.
\end{itemize}

Considering the unreliability of LLMs' internal states in hallucination detection, there are several promising alternative approaches, including prompting, probing and fine-tuning LLMs, which we explore in the next section.

\section{Prompting, Probing and Fine-Tuning} \label{sec:existing_methods}

Based on a review of the research area, we identify three groups of existing hallucination detection methods, which we discuss below.

\paragraph{Prompting} Prompting-based approaches involve directly prompting LLMs to assess the correctness of a given claim without additional training. We investigate the following three different methods: \Ni Prompting the model to output \emph{`True' or `False'} for a given claim, referred to as $\text{Prompt}_{\text{TF}}$. The probability assigned to the token `True' represents $P_{\text{factual}}$, while the probability assigned to `False' represents $P_{\text{hallucinated}}$. \Nii Prompting the model to output the \emph{probability} that it considers the given claim to be correct, referred to as $\text{Prompt}_{\text{Prob}}$. This number directly represents $P_{\text{factual}}$. \Niii SelfCheckGPT, which detects hallucinations by sampling additional responses from the model and assessing inconsistencies between each response and the target claim. The \emph{proportion} of responses that support the claim is taken as $P_{\text{factual}}$. 

\paragraph{Probing} Following \citet{su-etal-2024-unsupervised}, we train a multilayer perceptron (MLP) on the contextualized embeddings of LLMs to perform binary classification for hallucination detection, while keeping the base LLM frozen. The trained MLP outputs $P_{\text{factual}}$ as an indicator for classification.

\paragraph{Fine-Tuning} We fine-tune the base LLM with LoRA to enhance its ability to output `True' or `False' for a given claim \citep{kapoor2024large}. Similar to $\text{Prompt}_{\text{TF}}$, the probabilities assigned to the tokens `True' and `False' correspond to $P_{\text{factual}}$ and $P_{\text{hallucinated}}$, respectively. Note that LoRA fine-tuning allows us to easily use the original model for general tasks while applying the trained LoRA specifically for hallucination detection.

\begin{figure}[t]
  \centering
  \includegraphics[width=0.46\textwidth]{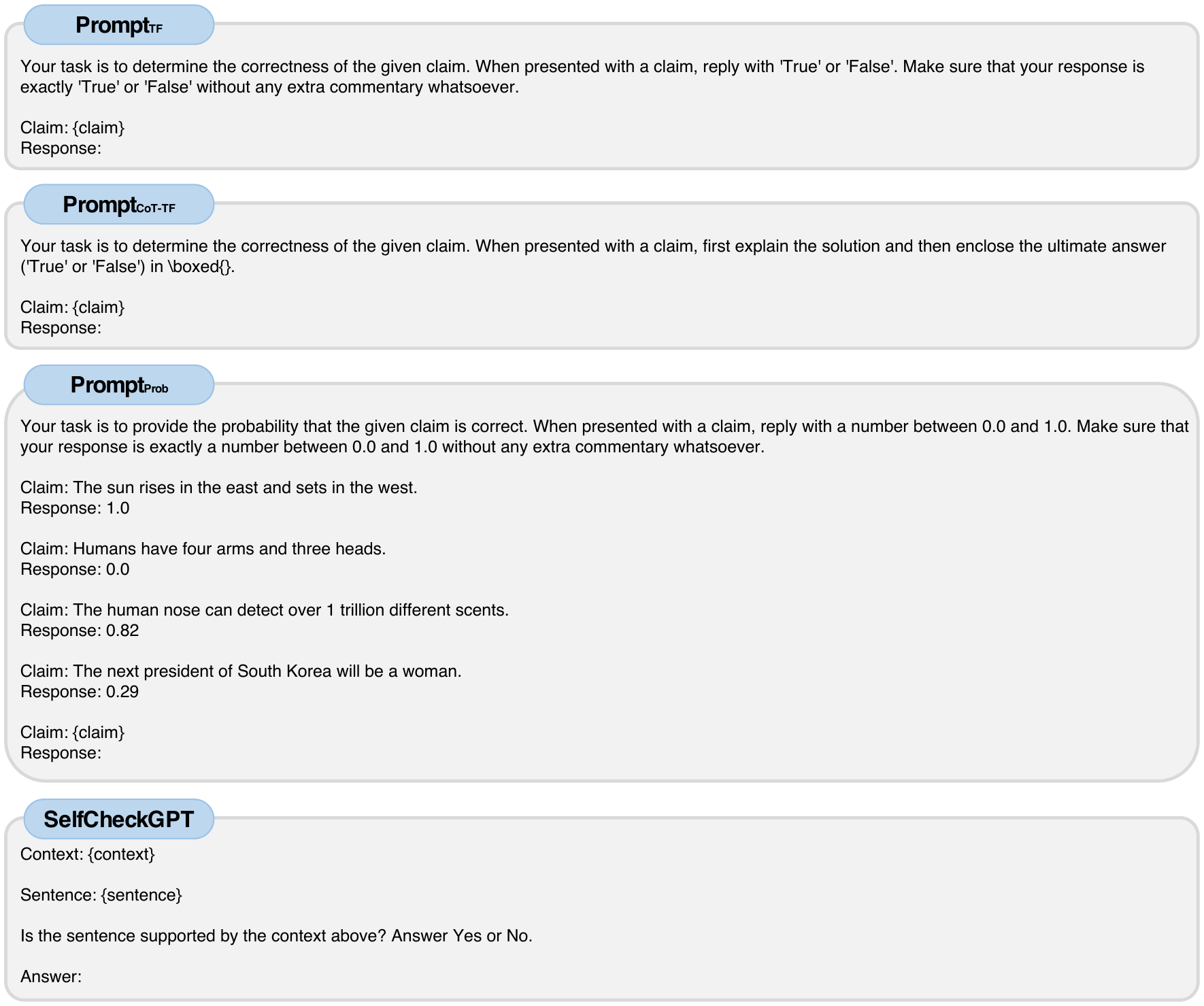}
    \captionof{figure}{Prompts for different prompting methods. $\text{Prompt}_{\text{CoT-TF}}$ refers to prompting the model to generate a reasoning path before making a judgment (detailed in Section \ref{sec:new_method_rateft}).}
    \label{fig:prompt_for_prompting}
\end{figure}

Figure \ref{fig:prompt_for_prompting} illustrates the prompts used for different prompting methods. The prompt used for constructing training data in Probing and Fine-Tuning is the same as the prompt employed by the $\text{Prompt}_{\text{TF}}$ method. For $\text{Prompt}_{\text{TF}}$ and $\text{Prompt}_{\text{Prob}}$, we obtain the response from the model with greedy decoding. Following \citet{manakul-etal-2023-selfcheckgpt}, we set the temperature to 1.0 and generate 20 additional responses for SelfCheckGPT. We evaluate 4 different types of contextualized embeddings for Probing: (1) the final token from the last layer ($\text{type}_{1}$), (2) the average of all tokens in the last layer ($\text{type}_{2}$), (3) the average of the final token across all layers ($\text{type}_{3}$), and (4) the average of $\text{type}_{1}$ and $\text{type}_{2}$ ($\text{type}_{4}$). The optimal embedding type, along with other hyperparameters, \eg\ learning rate, is selected through a search on the validation set. For Fine-Tuning and \rateft\ (introduced in Section \ref{sec:new_method_rateft}), we leverage the LLaMA-Factory library \citep{zheng-etal-2024-llamafactory} and perform a search on the validation set for important hyperparameters.

Following the data construction process outlined in Section \ref{sec:internal}, we conduct experiments on the full set of LongFact using Llama-3-8B-Instruct. This process yields 2,711 factual and hallucinated claims, which are subsequently split into training (70$\%$), validation (20$\%$), and test (10$\%$) sets. For all three types of methods, we use $P_{\text{factual}}$ as the classification indicator. Specifically, a claim is classified as `factual' if $P_{\text{factual}}$ exceeds a predefined threshold; otherwise, it is classified as `hallucinated'. The optimal threshold is determined through a search on the validation set. Consistent with \citet{tang2024minicheck,chen2024teaching}, we employ balanced accuracy (BAcc) as the evaluation metric: $\text{BAcc} = \frac{1}{2} (\frac{\text{TP}}{\text{TP+FN}} + \frac{\text{TN}}{\text{TN+FP}})$, where TP, TN, FP, and FN stand for true/false positives/negatives.

\begin{table}[t]    
    \centering
    \scalebox{0.60}{
    \begin{tabular}{lccccc}
        \toprule
        \multirow{2}{*}{\textbf{Dataset}} & \multicolumn{5}{c}{\textbf{Method}} \\
        \cmidrule(lr){2-6}
         & $\text{Prompt}_{\text{TF}}$  & $\text{Prompt}_{\text{Prob}}$ & SelfCheckGPT & Probing &  Fine-Tuning \\
        \midrule
        LongFact & 69.9 & 53.4  & 69.1 & 74.4 & \textbf{76.1} \\
        Biography & 72.3 &  56.3 & 71.9 & 77.0 & \textbf{78.2} \\
        \bottomrule
    \end{tabular}
    }
   \caption{
\label{tab:bacc_existing_methods}
BAcc ($\%$) of existing hallucination detection methods on LongFact and biography generation.
}
\end{table}

The results of different methods on the test set, as shown in Table \ref{tab:bacc_existing_methods}, indicate that fine-tuning LLMs is the most effective among all existing methods. While both $\text{Prompt}_{\text{TF}}$ and SelfCheckGPT achieve decent performance, Probing yields notable improvements by incorporating additional training with labels obtained from external search. Fine-Tuning further enhances performance by updating the internal features of LLMs, enabling more effective learning. In contrast, $\text{Prompt}_{\text{Prob}}$ performs significantly worse, likely due to LLMs' tendency to output high probabilities for hallucinated claims, leading to overconfidence. Additionally, we extend the experiments to biography generation \citep{min-etal-2023-factscore}. The results presented in Table \ref{tab:bacc_existing_methods} demonstrate that the observations and conclusions can be generalized to different datasets. We further verify the effectiveness of Fine-Tuning in Out-of-Distribution (OOD) scenarios by training the model on LongFact and evaluating its performance on Biography. The results reported in Table \ref{tab:ood_results} demonstrate that Fine-Tuning effectively generalizes to OOD scenarios.

\begin{table}[t]    
    \centering
    \scalebox{0.71}{
    \begin{tabular}{ccccc}
        \toprule
        $\text{Prompt}_{\text{TF}}$ & $\text{Prompt}_{\text{Prob}}$ & SelfCheckGPT & Probing & Fine-Tuning  \\
        \midrule
        72.3 & 56.3 & 71.9 & 71.1 & \textbf{74.7} \\
        \bottomrule
    \end{tabular}
    }
   \caption{
\label{tab:ood_results}
Performance of different methods in Out-of-Distribution (OOD) scenarios, where Fine-Tuning continues to achieve the best results.
}
\end{table}

Building on these findings, a natural question arises: can Fine-Tuning be further improved to develop more effective hallucination detectors? We answer this question by \emph{incorporating rationales and an auxiliary task into the training process.}

\section{Rationale and Auxiliary Task Enhanced Fine-Tuning (\rateft)} \label{sec:new_method_rateft}

While hallucination detection is not regarded as a reasoning task in the conventional sense, incorporating Chain-of-Thought (CoT) \citep{wei2022chain} explaining the judgment can still be beneficial for distinguishing factual content from hallucinated information as it enables LLMs to better evaluate the correctness of claims by systematically analyzing underlying components. To examine the impact of rationales, we prompt the model to generate a reasoning path before making a judgment (\ie\ `True' or `False'), referred to as $\text{Prompt}_{\text{CoT-TF}}$. This approach improves performance from 69.9 (using $\text{Prompt}_{\text{TF}}$) to 74.9, highlighting the effectiveness of incorporating CoT reasoning.

\paragraph{Augmenting Fine-Tuning with Rationales} Building on the above observation, 
we augment the fine-tuning dataset with rationales generated by the model during data construction, explaining whether the search results support the claims. Notably, we adopt the `label-rationale' format to maintain the same inference cost as the baseline Fine-Tuning. This allows us to directly derive $P_{\text{factual}}$ from the first output token without requiring the generation of the complete reasoning path.

Consolidating knowledge through repetition in diverse contexts is a fundamental principle of effective human learning \citep{ausubel2012acquisition}. For example, medical students deepen their understanding of anatomy by studying diagrams, practicing in simulations, and engaging in hands-on dissections, each offering a unique perspective on the same foundational knowledge. Drawing inspiration from this paradigm, we introduce an auxiliary question answering (QA) task into the fine-tuning process to further strengthen the model's understanding and enhance its generalization capabilities. This auxiliary QA task serves as a complementary component to the primary hallucination detection task, offering the model an alternative but closely related perspective on the problem (see Section \ref{sec:more_analysis_auxiliary_task} for more analysis on the auxiliary task).

\paragraph{Augmenting Fine-Tuning with QA Task} Specifically, for each claim, we first prompt the model to generate a question about the key information within it. If the claim is factual, we ask the model to extract the correct answer directly from the claim and provide an explanation, forming a QA example. For hallucinated claims, we leverage the augmented rationale to guide the model in generating an appropriate correct answer along with an explanation. After constructing these QA examples, they are combined with the original data for fine-tuning.

\begin{table}[t]
\centering
    \scalebox{0.60}{
    \begin{tabular}{lccccc}
    \toprule
    \multirow{2}{*}{\textbf{Dataset}} & \multicolumn{5}{c}{\textbf{Method}} \\
    \cmidrule(lr){2-6}
    & $\text{Prompt}_{\text{TF}}$ & $\text{Prompt}_{\text{CoT-TF}}$ & Probing & Fine-Tuning & \rateft \\
    \midrule
    LongFact & 69.9 & 74.9 & 74.4 & 76.1 & \textbf{79.6} \\
    Biography & 72.3 & 74.8 & 77.0 & 78.2 & \textbf{80.9} \\
    \bottomrule
    \end{tabular}
    }
\caption{
\label{tab:comparison_rate_ft_others}
{
BAcc ($\%$) of \rateft\ and baseline methods. \rateft\ is significantly better than Fine-Tuning with $p$-value $<0.01$.
} 
}
\end{table}

\begin{figure}[t]
  \centering
  \includegraphics[width=0.46\textwidth]{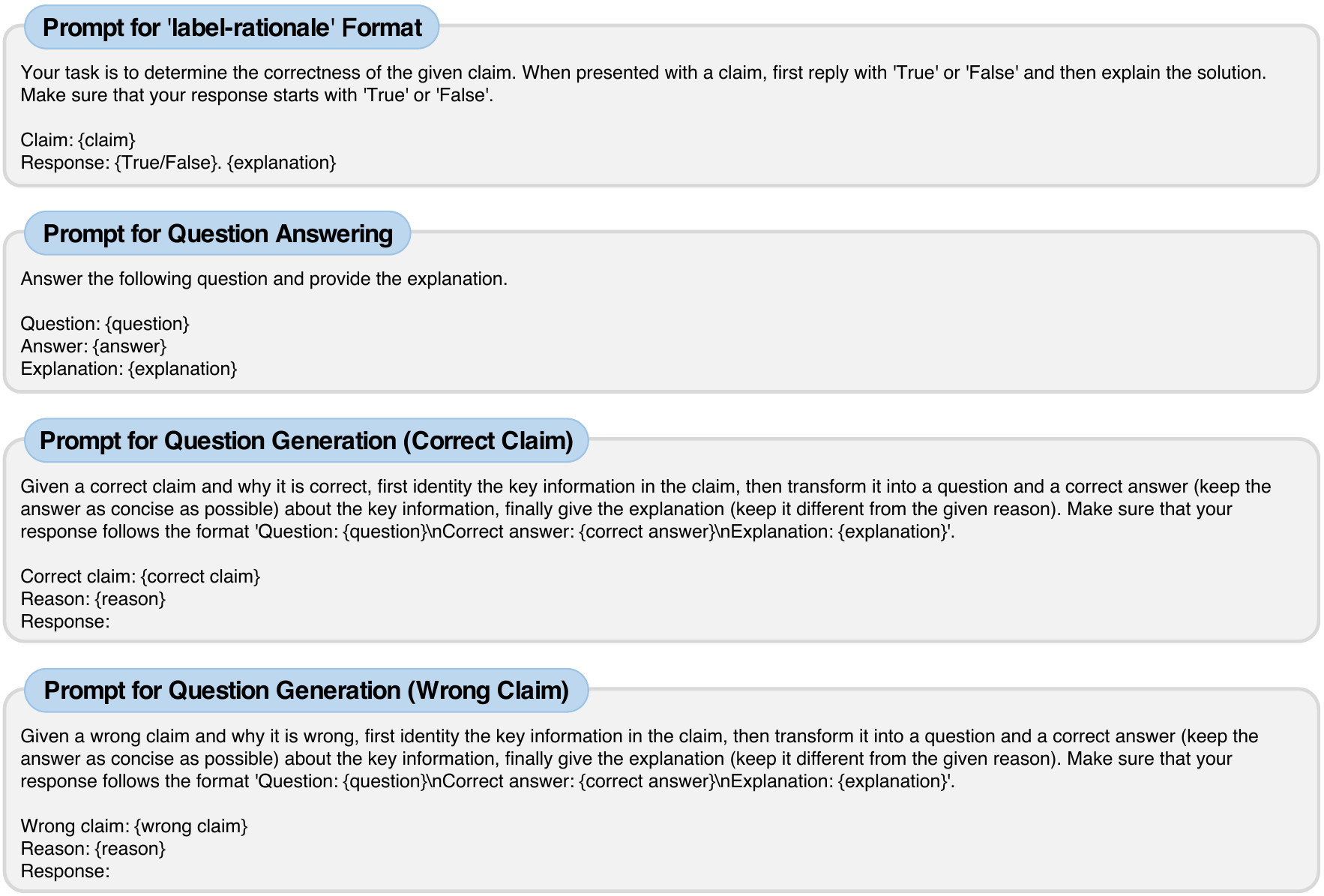}
    \captionof{figure}{Prompts used for different components of \rateft.}
    \label{fig:prompt_for_rateft}
\end{figure}

By integrating these two strategies, we propose \textbf{R}ationale and \textbf{A}uxiliary \textbf{T}ask \textbf{E}nhanced \textbf{F}ine-\textbf{T}uning (\rateft) (Figure \ref{fig:method_rateft}). \rateft\ requires the model to systematically analyze and explain its judgments and allows the model to benefit from complementary learning perspectives, reinforcing its understanding of claims through diverse yet interconnected tasks (see Figure \ref{fig:prompt_for_rateft} for all the prompts used in \rateft). Following the experimental setup described in Section \ref{sec:existing_methods}, we show the comparison between \rateft\ and baseline approaches in Table \ref{tab:comparison_rate_ft_others}, which demonstrates the superiority of \rateft\ across different datasets.

To isolate the effect of additional data augmentation versus the auxiliary QA task, we design two variants: \Ni we paraphrase the original claim using GPT-4 for data augmentation and fine-tune the model on the combined data, referred to as $\text{Fine-Tuning}_{\text{para}}$, which has roughly the same amount of training data as \rateft; and \Nii we reduce the training data for \rateft\ by half (approximately the same amount as Fine-Tuning), referred to as $\text{RATE-FT}_{\text{half}}$. We conduct experiments on LongFact using Llama-3-8B-Instruct and present the results in Table \ref{tab:comparison_ftpara_rateft} and \ref{tab:comparison_ft_ratefthalf}, which demonstrate that the performance improvement primarily comes from our designed auxiliary task, rather than from additional data augmentation.

\begin{table}[t]    
    \centering
    \scalebox{0.75}{
    \begin{tabular}{cc}
        \toprule
        $\text{Fine-Tuning}_{\text{para}}$ & \rateft \\
        \midrule
         76.8 & \textbf{79.6} \\
        \bottomrule
    \end{tabular}
    }
   \caption{
\label{tab:comparison_ftpara_rateft}
Comparison between $\text{Fine-Tuning}_{\text{para}}$ and \rateft.
}
\end{table}

\begin{table}[t]    
    \centering
    \scalebox{0.75}{
    \begin{tabular}{cc}
        \toprule
        Fine-Tuning & $\text{RATE-FT}_{\text{half}}$ \\
        \midrule
        76.1 & \textbf{78.5} \\
        \bottomrule
    \end{tabular}
    }
   \caption{
\label{tab:comparison_ft_ratefthalf}
Comparison between Fine-Tuning and $\text{RATE-FT}_{\text{half}}$.
}
\end{table}

\subsection{Further Analysis}

\begin{table}[t]
\centering
    \scalebox{0.70}{
    \begin{tabular}{lcccc}
    \toprule
    \multirow{2}{*}{\textbf{Dataset}} & \multicolumn{4}{c}{\textbf{Method}} \\
    \cmidrule(lr){2-5}
    & Fine-Tuning  & \emph{w.o.} aux & \emph{w.o.} rationale & \rateft \\
    \midrule
    LongFact & 76.1 & 77.5 & 77.9 & \textbf{79.6} \\
    Biography & 78.2 & 79.4 & 79.5 & \textbf{80.9} \\
    \bottomrule
    \end{tabular}
    }
\caption{
\label{tab:ablation_study_res} {Results of different ablations. Removing the auxiliary task or rationales leads to a noticeable performance drop.} 
}
\end{table}

\subsubsection{Ablation Study} \label{sec:ablation_study}

We analyze the contribution of different components of \rateft\ by investigating two variants of \rateft\: \Na without the auxiliary task (\emph{w.o.} aux) and \Nb without rationales (\emph{w.o.} rationale). Table \ref{tab:ablation_study_res} presents the performance of different methods, highlighting that each component plays an important role in achieving the overall performance.

\begin{table}[t]
\centering
    \scalebox{0.52}{
    \begin{tabular}{lccccc}
    \toprule
    \multirow{2}{*}{\textbf{Model}} & \multicolumn{5}{c}{\textbf{Method}} \\
    \cmidrule(lr){2-6}
    & $\text{Prompt}_{\text{TF}}$ & $\text{Prompt}_{\text{CoT-TF}}$ & Probing & Fine-Tuning & \rateft \\
    \midrule
    Llama-3.1-70B-Instruct & 73.2 & 76.8 & 79.4 & 80.6 & \textbf{83.8} \\
    Mistral-7B-Instruct & 61.8 & 64.1 & 68.4 & 70.8 & \textbf{73.4} \\
    Qwen2.5-7B-Instruct & 72.8 & 75.5 & 77.0 & 78.4 & \textbf{81.1} \\
    \bottomrule
    \end{tabular}
    }
\caption{
\label{tab:generalization_diff_models}
{
BAcc ($\%$) of \rateft\ and baselines using different models. \rateft\ consistently outperforms baseline methods across all models.
} 
}
\end{table}

\subsubsection{Generalization to Different Models} 

Our experiments and analysis so far use Llama-3-8B-Instruct as the backbone model. To verify whether the performance gain of \rateft\ is consistent across different backbone models, we extend the experiments to Llama-3.1-70B-Instruct \citep{dubey2024llama}, Mistral-7B-Instruct \citep{jiang2023mistral}, and Qwen2.5-7B-Instruct \citep{yang2024qwen2} on LongFact. When using other models, we follow the exact same settings as those used for Llama-3-8B-Instruct. Therefore, our constructed benchmarks align well with \citet{su-etal-2024-unsupervised}, as both include responses and internal states from various LLMs. The key difference is that the LLMs we investigate are all modern models, whereas the models used in \citet{su-etal-2024-unsupervised} are relatively outdated (such as LLaMA-2 and GPT-J).

From the results shown in Table \ref{tab:generalization_diff_models}, we can observe that \rateft\ consistently outperforms baseline approaches across all models, demonstrating its robustness and generalizability to diverse model architectures and scales.

\subsubsection{More Analysis on Auxiliary Task} \label{sec:more_analysis_auxiliary_task}

\paragraph{Comparison with F2} F2 \citep{hu2024mitigating} also integrates rationales and auxiliary tasks into the training process. However, its main goal is to enhance the faithfulness of model responses while we focus on improving the accuracy of hallucination detection.

\paragraph{Further Clarification on Motivation} The underlying motivation for introducing the auxiliary question answering (QA) task into fine-tuning is that hallucination detection and mitigation are complementary and closely related tasks. This auxiliary QA task—where a question about the key information in the claim is posed, and the model is trained to provide the correct answer—helps improve the factuality of the model’s responses through supervised fine-tuning. It acts as a complementary component to the primary hallucination detection task, offering the model an alternative yet closely related perspective, thereby enhancing its generalization capabilities.

\begin{table}[t]    
    \centering
    \scalebox{0.75}{
    \begin{tabular}{cccc}
        \toprule
         $\text{Prompt}_{\text{CoT-TF}}$ & Probing & Fine-Tuning & \rateft \\
        \midrule
        80.4 & 81.1 & 82.4 & \textbf{85.0} \\
        \bottomrule
    \end{tabular}
    }
   \caption{
\label{tab:bacc_unknown_longfact}
BAcc-unknown ($\%$) of different methods on Longfact with Llama-3-8B-Instruct.
}
\end{table}

\subsubsection{Incorporating Uncertainty for Hallucination Detection} \label{sec:uncertainty_for_detection}

To enhance hallucination detection, we propose incorporating model uncertainty into the detection process, enabling a hybrid pipeline that combines the strengths of the model and external tools. Specifically, when the model is uncertain about whether a claim is factual or hallucinated, we leverage external tools to handle ambiguous cases, improving overall performance. The process involves setting two thresholds, $\alpha_{low}$ and $\alpha_{high}$, for classification. A claim is classified as `factual' if $P_{\text{factual}} > \alpha_{high}$ and `hallucinated' if $P_{\text{factual}} < \alpha_{low}$. Claims falling between these thresholds are classified as `unknown' and delegated to external tools for further evaluation. Assuming the external tools' output is the ground truth, predictions classified as `unknown' are treated as correct. To evaluate the hybrid pipeline, we define the BAcc-unknown metric as follows:  

\small
\begin{equation}
\begin{aligned}
\text{BAcc-unknown} &= \frac{1}{2} ( \frac{\text{\# Correct Factual Predictions}}{\text{\# Total Factual Claims}} \\
& + \frac{\text{\# Correct Hallucinated Predictions}}{\text{\# Total Hallucinated Claims}} )
\end{aligned}
\end{equation}
\normalsize
The optimal thresholds, $\alpha_{low}$ and $\alpha_{high}$, are determined through a search on the validation set. This process ensures that BAcc on the validation set exceeds 70$\%$, while also maximizing BAcc-unknown. The goal is to strike a balance between performance and efficiency by achieving high BAcc-unknown without generating an excessive number of `unknown' predictions, which could substantially increase detection costs. We conduct experiments on LongFact using Llama-3-8B-Instruct and report the results in Table \ref{tab:bacc_unknown_longfact}, which demonstrate that incorporating model uncertainty greatly enhances hallucination detection, as evidenced by the BAcc-unknown metric's superior performance compared to standard BAcc in resolving ambiguous cases. Moreover, \rateft\ continues to outperform all other methods with respect to the BAcc-unknown metric, highlighting its robustness and effectiveness.

\begin{table}[t]    
    \centering
    \scalebox{0.68}{
    \begin{tabular}{lccc}
        \toprule
          & length $<$ 500 & 500 $\le$ length $\le$ 1000 & length $>$ 1000 \\
        \midrule
        Fine-Tuning & 74.4 & 77.2 & 75.1 \\
        \rateft & \textbf{77.6} & \textbf{80.9} & \textbf{78.0} \\
        \bottomrule
    \end{tabular}
    }
   \caption{
\label{tab:scale_response_length}
The performance of Fine-Tuning and \rateft\ with different response lengths.
}
\end{table}

\subsubsection{Scaling with Response Length} 

To evaluate how the performance of baselines and \rateft\ scales with response length, we divide the LongFact dataset (with responses generated by Llama-3-8B-Instruct) into three subsets: responses under 500 tokens, between 500 and 1000 tokens, and over 1000 tokens. Using the same experimental setup as the main study, we compare Fine-Tuning with \rateft\ across these subsets. The results reported in Table \ref{tab:scale_response_length} show that \rateft\ can consistently outperform Fine-Tuning with different response lengths.

\subsubsection{Comment on Reliance on External Search} 

While the data construction process for fine-tuning relies on external search engines like Google for annotation, \rateft\ is \emph{fully reference-free at inference time}—the use of external search is limited to a one-time offline process for constructing the benchmark and obtaining supervision signals. Once trained, the detector no longer depends on any external tools or APIs. This design choice ensures that \rateft\ remains practical and easily deployable in real-world scenarios where external fact-checking tools are unavailable or costly.

\subsubsection{Detecting Faithfulness Hallucinations} 

As outlined in Section \ref{sec:related_work}, hallucinations are typically categorized based on what they contradict: factuality hallucinations, which conflict with real-world knowledge, and faithfulness hallucinations, which contradict the input context, instruction, or internal logic. This work primarily focuses on \emph{factuality hallucinations}, aiming to develop a reference-free approach capable of detecting whether generated content contradicts real-world facts. Faithfulness hallucinations, such as inconsistencies with the input context, are not the focus of this study. They generally do not require external fact-checking tools for verification and are thus beyond the scope of our current method.

In addition, we show the output distribution of the fine-tuned model, the comparison between \rateft\ and methods using external fact-checking tools, and the prompts used for output extraction in Appendix \ref{sec:output_distribution_finetuned_model} $\sim$ \ref{sec:prompt_claim_extraction}, respectively.

\section{Conclusion}

In this work, we systematically investigate reference-free hallucination detection in open-domain long-form generation. Our study begins with an analysis of the model's internal states, demonstrating that these states alone cannot reliably detect hallucinations. We then evaluate several existing approaches, including prompting, probing, and fine-tuning, with fine-tuning emerging as the most effective method. Building on these findings, we introduce Rationale and Auxiliary Task Enhanced Fine-Tuning (\rateft), a novel approach that leverages rationales and an auxiliary task to achieve significant improvements in detection performance across two datasets and various LLMs.

\section*{Limitations}

One limitation of our work is its focus solely on improving the performance of the hallucination detector. A potential improvement could be to explore leveraging the detector's feedback as a reward signal to guide LLMs to generate more factual responses. Additionally, developing a more comprehensive benchmark for hallucination detection in open-domain long-form generation that covers a broader range of domains would further enhance its applicability.

\bibliography{anthology,custom}

\appendix

\section{Appendix} \label{sec:appendix}

\begin{figure}[t]
  \centering
  \includegraphics[width=0.46\textwidth]{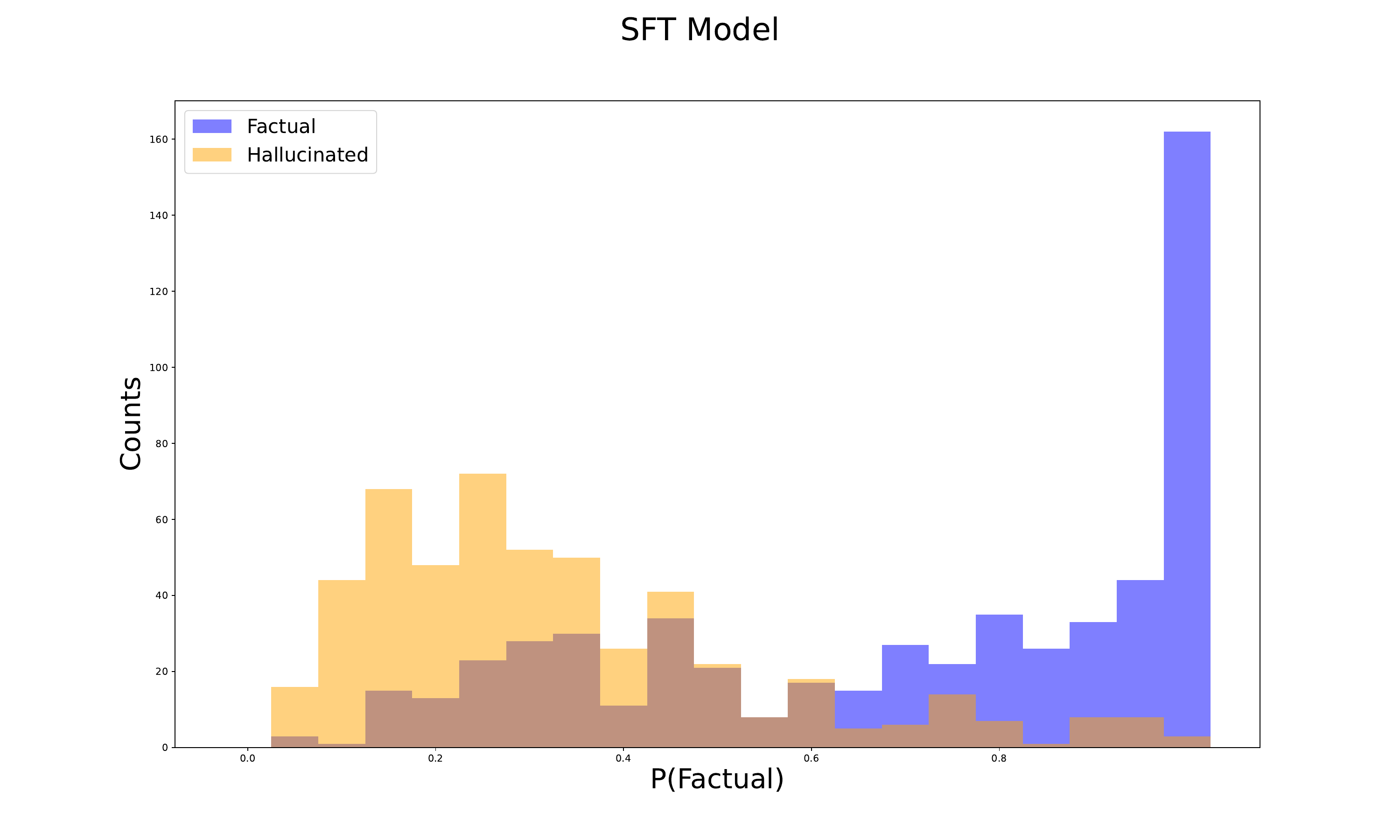}
    \captionof{figure}{Model's $P_{\text{factual}}$ after applying \rateft\ for both factual and hallucinated claims.}
    \label{fig:distribution_finetuned_model}
\end{figure}

\subsection{Output Distribution of Fine-Tuned Model} \label{sec:output_distribution_finetuned_model}

We present the distribution of the model’s $P_{\text{factual}}$ after applying \rateft\ for both factual and hallucinated claims in Figure~\ref{fig:distribution_finetuned_model}. Compared to the internal state-based results shown in Figures~\ref{fig:token_probability} to~\ref{fig:token_entropy_entity}, \rateft\ leads to a significantly improved separation between factual and hallucinated claims, demonstrating its effectiveness in enhancing the model's discriminative capability.

\subsection{Comparison with Methods Using External Fact-Checking Tools} \label{sec:comparison_using_external_factchecking}

Methods that rely on external fact-checking tools can be viewed as an upper bound for RATE-FT. As shown in Tables \ref{tab:comparison_rate_ft_others} and \ref{tab:generalization_diff_models}, RATE-FT achieves over 80$\%$ of the performance compared to using ground-truth annotations obtained from Google Search, demonstrating a favorable trade-off between effectiveness and independence from external resources.

Additionally, Section \ref{sec:uncertainty_for_detection} illustrates how RATE-FT can be integrated with external tools in high-uncertainty cases. This highlights the flexibility of our approach and its potential to further improve performance when external tools are available.

\begin{figure}[t]
  \centering
  \includegraphics[width=0.46\textwidth]{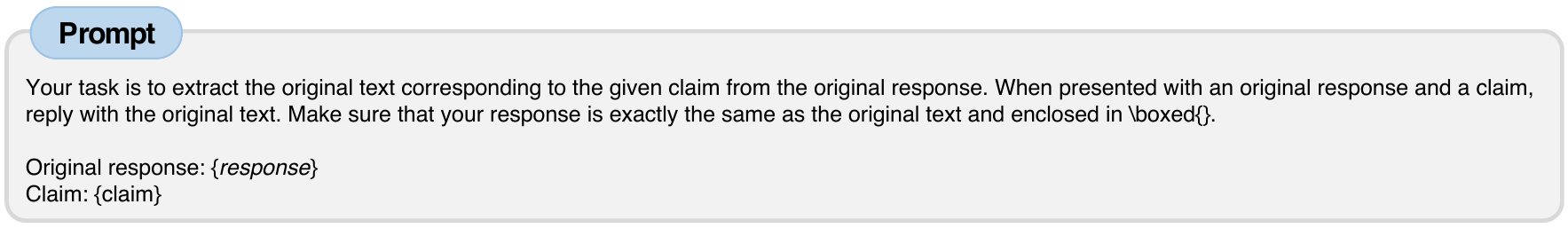}
    \captionof{figure}{Prompt for extracting the original output given an atomized claim.}
    \label{fig:prompt_extraction}
\end{figure}

\subsection{Prompt for Output Extraction} \label{sec:prompt_claim_extraction}

After decomposition, the atomized claims may differ from the original expression in the response. To address this, we use the prompt shown in Figure \ref{fig:prompt_extraction} to retrieve the original output corresponding to a given atomized claim.

\end{document}